\documentclass{article}
\usepackage{spconf,amsmath,graphicx}
\usepackage{multicol,multirow}
\usepackage{changes}

\usepackage{xcolor}

\definecolor{darkspringgreen}{rgb}{0.09, 0.7, 0.27}

\title{Deliberation of Streaming RNN-Transducer \\ by  Non-autoregressive Decoding}
\name{Weiran Wang $\qquad$ Ke Hu $\qquad$ Tara N. Sainath}
\address{Google, Inc.\\
\{weiranwang, huk, tsainath\}@google.com}
\begin{document}
\ninept
\maketitle
\begin{abstract}
We propose to deliberate the hypothesis alignment of a streaming RNN-T model with the previously proposed Align-Refine non-autoregressive decoding method and its improved versions. The method performs a few refinement steps, where each step shares a transformer decoder that attends to both text features (extracted from alignments) and audio features, and outputs complete updated alignments. The transformer decoder is trained with the CTC loss which facilitates parallel greedy decoding, and performs full-context attention to capture label dependencies. We improve Align-Refine by introducing cascaded encoder that captures more audio context before refinement, and alignment augmentation which enforces learning label dependency.
We show that, conditioned on hypothesis alignments of a streaming RNN-T model, our method obtains significantly more accurate recognition results than the first-pass RNN-T, with only small amount of model parameters. 
\end{abstract}
\section{Introduction}
\label{sec:intro}




Deliberation is a two-pass modeling paradigm where a second-pass model is employed to refine decoding results of a first-pass model~\cite{xia2017deliberation}, by either re-decoding or rescoring first-pass hypotheses.
Both first-pass hypotheses and ground truth pairs are presented in a supervised way for training the second-pass model.
Previously, second-pass LAS models based on long-short term memory (LSTM) networks~\cite{hu2020deliberation} or transformer decoder~\cite{hu2021transformer} were shown to significantly improve Google voice search quality over the first-pass and an acoustic-rescoring model~\cite{sainath2020streaming}.
One potential disadvantage of prior deliberation methods, however, is that the second-pass decoder still works in an autoregressive manner which can be slow.

On the other hand, there has been a recent surge of interests on \emph{non-autoregressive} sequence models that are not constrained to decode in the left-to-right fashion~\cite{ghazvininejad2019MaskPredict,chen2019listen,higuchi2020mask,higuchi2021improved,wang2021streaming,chan20,chi2020align}. These models make parallel update steps during decoding, i.e., each decoding step can modify multiple or all positions of previous decoding results simultaneously. And in order to gradually capture sufficient label dependency, these models usually require multiple decoding steps to achieve good performance.
While parallel decoding methods are faster, empirically it is challenging for a purely non-autoregressive model to match the accuracy of a (similarly-sized) single-pass autoregressive model, which captures the dependency explicitly in its learning objective and the decoding process.

In this paper, we propose to deliberate a small streaming RNN-T model with non-autoregressive decoding, to achieve the best of both worlds. The motivation is that a small autoregressive model can produce hypotheses of good word error rate (WER) with low latency, and the second-pass can take into account label dependency with right context to further improve accuracy, while facilitating a simple and efficient decoding method (e.g., parallel greedy decoding) during deliberation. This approach allows us to trade off latency, accuracy, model size, and perhaps on-device energy consumption.

We design our deliberation method based on the previously proposed Align-Refine algorithm~\cite{chi2020align}. This algorithm refines the hypothesis alignment 
(i.e., the sequence of discrete tokens corresponding to inferred labels at each frame) from the first pass in a few steps, where each step outputs complete alignment conditioned on the input one. Align-Refine is trained with the CTC loss and thus facilitates parallel decoding, yet it builds label dependency implicitly into the model prediction through attention operations on the alignments. 
We propose improvements to Align-Refine to further capture wider audio context in the deliberation setup, and to enforce label dependency modeling through alignment augmentation.
Experiment results on voice search show that we can improve  the WER of a 56M first-pass RNN-T model from 7.8\% to 6.4\% using a 30M Align-Refine model, and our improved version reduces the WER further to 5.7\% with additional 25M parameters, and most of the gain can be achieved in 2 refinement steps with parallel greedy decoding.

\section{Related work}
\label{sec:format}



The non-autoregressive decoding literature focuses on developing models and decoding algorithms that are not constrained to work in the left-to-right fashion, in which every output token is decoded based on the history of previous tokens (e.g., beam search). End-to-end models that explicitly capture label dependency (e.g., LAS~\cite{Chan_16a} and RNN-Transducer~\cite{Graves_12a}) are in general autoregressive, while models that make label independence assumptions (e.g., CTC~\cite{graves2006connectionist}) facilitate parallel decoding of all tokens at the same time (although beam search may still improve accuracy). 

Non-autoregressive models can be divided into two categories. The first category of models work on the label sequences~\cite{chen2019listen,higuchi2020mask,higuchi2021improved,wang2021streaming}. This approach generally follows the mask-predict paradigm~\cite{ghazvininejad2019MaskPredict}, where in each step certain positions (e.g., the least confident predictions) are masked (replaced by a special \texttt{[mask]} token) in the input label sequence. Based on the masked input, the model predicts all output labels simultaneously, and replace the input masks with the predictions in those positions. A challenge to this approach is to estimate the length of the label sequence.
The second category of methods work on alignments instead, which are sequences containing the underlying token for each frame. An early work of this category is Imputer~\cite{chan20}. Starting from all masked positions, Imputer divides the alignment into blocks, and recovers the final alignment in a fixed number of steps, where in each step a single position of each block is predicted. During training, the model takes randomly masked ground truth alignments or forced alignments as input, and predicts the masked positions; the training loss is a variant of the CTC loss that only measures loss on masked positions. It is easy to see that this approach suffers from exposure bias, as the method is not exposed to decoding errors from the model itself during training. 
Our deliberation method is based on the Align-Refine algorithm of~\cite{chi2020align} which also works on alignment but does not suffer from the exposure bias; we review it in Sec~\ref{sec:Align-Refine}.

\section{Non-autoregressive deliberation}
\label{sec:method}

\subsection{First-pass modeling by RNN-T}
\label{sec:rnnt}

We use a streaming RNN-T model~\cite{Graves_12a, He_18a} to generate first-pass hypothesis. The 
model has a causal encoder $enc_0$ that extracts audio features from the input utterances $X$, denoted by $enc_0(X)$ with length (number of frames) $T'$. The decoder, 
denoted by $dec_0$, consists of the prediction network for modeling label dependency similarly to a language model, and the joint networks for combining audio and language model features and outputting per-frame posterior of labels. We perform beam search with $dec_0$ on top of $enc_0(X)$, which reasons about the hidden label at each frame based on the posterior, and feeds the inferred label back to the prediction network if it is non-blank. Beam search returns the \emph{alignments} of probable hypotheses for an utterance: each alignment is a sequence of discrete tokens corresponding to the inferred labels (possibly blanks) at each frame, and discarding the blanks (indicating non-emitting frames) of an alignment reduces it to the final hypothesis. Denote the alignment of consideration (e.g., the alignment for the 1-best hypothesis) from RNN-T by $A^0 (X)$ with length $T$. Note $T$ may be greater than $T'$, as RNN-T can output multiple labels at a frame.

\subsection{Incorporating Iterative Non-autoregressive Refinement}
\label{sec:Align-Refine}

We now review the Align-Refine method~\cite{chi2020align} and describe how it is used in our deliberation setup.
The first-pass hypothesis from RNN-T is then fed to the Align-Refine decoder, denoted as $dec_1$, for $R$ steps of iterative refinement. For the $i$-th refinement step, $dec_1$ takes in an initial alignment $A^{i-1}$ and output a new alignment $A^{i}$; all steps ($i=1,\dots,R$) share the same model parameters of $dec_1$.

Similarly to the decoder module in attention-based model~\cite{Vaswani_17a}, $dec_1$ integrates the text-side information and the audio-side information, with a series of transformer layers. For the $i$-th refinement step, the input alignment $A^{i-1}$ is first mapped to continuous embedding vectors, which serve as the initial text-side features. Each transformer layers performs a self-attention for the text features, and then use its result (of length $T$) as query to perform cross-attention on $enc_0(X)$ where audio features are used as both the key and value. The output of each transformer layer (of length $T$) is used as the text-side features for the next layer. A final softmax layer is used on top of the last transformer layer output to predict real labels and blanks.

The greedy alignment from CTC, computed by picking the most probably token for each of the $T$ positions (this can be done for all frames in parallel), is used as the output alignment for the $i$-th refinement step and input to the $(i+1)$-th step. 
We train all layers in a refinement step jointly with the CTC loss~\cite{graves2006connectionist}, which marginalizes all alignments for the ground truth label sequence. 
The overall training objective is the average of CTC losses of all $R$ steps.
During inference, at the end of the $R$-th refinement steps, we collapse the output alignment $A^R$ into a label sequence with the operator $\mathcal{B}$ which removes repetitions and then blanks, to yield the predicted label sequence  $\mathcal{B}(A^R)$.

To summarize, the main differences between the Align-Refine decoder and a regular attention decoder are twofold.
    First, the regular attention decoder takes the label sequence as text-side input, whereas the Align-Refine decoder takes the alignment as input. For the former, the output label sequence is aligned with the input label sequence so a simple cross-entropy loss is used for training. For the latter, as alignment is typically longer than label sequence, one has to reason about the time correspondence between positions of the alignment and the output labels, for which the CTC loss is a natural choice. On the other hand, if we look at the collapsed label sequence for the alignments $\mathcal{B}(A^1), \dots,  \mathcal{B}(A^R)$, the refinement steps potentially allow complicated edits (insertion, deletion, substitution) from initial RNN-T hypotheses.
    This is in contrast to the Imputer approach~\cite{chan20} which gradually removes masks over the alignment, not allowing modifications to the already revealed positions, and different from early mask-predict approaches~\cite{chen2019listen} where the hypothesized label sequence length can not be changed over iterations.
    Second, the regular attention decoder trains and decodes in the autoregressive fashion and though only past label history is considered for predicting the current label, whereas the Align-Refine decoder employs full-context attention to capture also ``future" label information. Therefore, rich label context is already built into the model predictions and parallel decoding works well, alleviating us from repeatedly conditioning on the past as in autoregressive decoding.

\begin{figure}[t]
    \centering
    \includegraphics[width=0.9\linewidth]{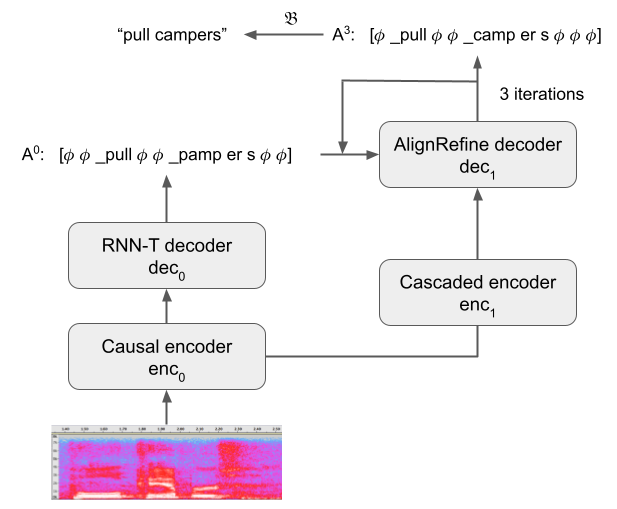}
    \vspace*{-1em}
    \caption{Schematic diagram of Align-Refine with cascaded encoder. $\phi$ denotes blank for RNN-T and CTC in alignments.} 
    \label{fig:Align-Refine}
\end{figure}

\subsection{Improvements to Align-Refine}
\label{sec:improvements}

\textbf{Additional audio feature processing by cascaded encoder} The original Align-Refine decoder uses self-attention on the text-side input to capture label dependency, and uses cross-attention to model the alignment between label and audio features; no further processing was done on the audio features during refinement steps. As an improvement to the original formulation, we propose to extract audio features of rich (right) context and use them as input to Align-Refine, instead of the causal encoder output. To this end, we introduce an additional cascaded encoder~\cite{arun21cascade}, denoted by $enc_1$, between the causal encoder $enc_0$ and the deliberation decoder $dec_1$. Cascaded encoder was originally introduced to unify streaming and non-streaming ASR models, and in our work $enc_1$ consists of a few conformer layers~\cite{gulati2020conformer} which perform self-attentions on top of $enc_0(X)$ with certain right context. Figure~\ref{fig:Align-Refine} illustrates the overall method with cascaded encoder.

Empirically, $enc_1$ helps to improve the accuracy of Align-Refine significantly, compared to using $enc_0$ alone, demonstrating the importance of audio feature quality in non-autoregressive decoding.
Our approach is appropriate when we have strict latency constraint for the first-pass model that precludes the use of right context, but can tolerate some delay for the second-pass model which takes advantage of parallel computations. 

\noindent\textbf{Alignment augmentation} To enforce the refinement steps to capture more label dependencies, we can perform augmentation on the alignments $A^0, \dots, A^R$. Here we employ a common augmentation technique used for text data (also represented as a sequence of discrete tokens) known as masking~\cite{devlin2018bert}: we introduce noise into the alignment by randomly replacing each label (including blanks) with a special \texttt{[mask]} token (not in the output vocabulary) for some probability. During training, the model must make the correct final prediction in presence of \texttt{[mask]}, leveraging information from both audio and text inputs. We find small masking probabilities leads to small but consistent gain. Note that unlike the mask-predict approach where masks are an integral part of the prediction procedure, the masks we introduce here are purely for the purpose of augmentation and are not needed for inference.

On the other hand, we apply SpecAugment~\cite{park2019specaugment} on the audio data when forwarding the first-pass model, which may cause additional decoding errors compared to using clean spectrogram features. Thus SpecAugment already provides an indirect form of alignment augmentation. We have also tried to include not only the top alignment from RNN-T but the top-k with $k=2$ and $4$ for training (and still refine the top-1 alignment during inference), but this form of augmentation barely improved the final performance.
It is future work to explore other forms of augmentation in the literature (e.g., the shifting augmentation from~\cite{chan20}).

\section{Experimental results}
\label{sec:expts}


\subsection{Setup}
\label{sec:setup}

We use the same dataset of~\cite{narayanan2019recognizing} for training, including anonymized and hand-transcribed audio data covering the search, farfield, telephony and YouTube domains. The audio has gone through multi-condition training (MTR~\cite{kim2017generation}) and random 8kHz down-sampling \cite{li2012improving} to increase diversity.

The development set for hyper-parameters tuning, denoted by VS,  contains around 12K anonymized and hand-transcribed utterances that are representative of Google’s Voice Search traffic, with an average duration of 5.5 seconds.
We report final WERs on five test sets. The first test set is side-by-side losses (SXS) set, which contains a set of 1K utterances where the quality of the E2E model
transcription has more errors than a state-of-the-art conventional model~\cite{Golan16}. 
The other four are TTS generated test sets containing rare proper nouns (RPN) which appear less than 5 times in the training set. These sets cover the Maps, News, Play, and QSession domains and are denoted RPN-M, RPN-N, RPN-P, and RPN-Q respectively, each containing 10K utterances.


\subsection{Model specifications}
\vspace*{-.5em}

Inputs to our models are 128-dimensional log-Mel features, computed with a 32ms window and shifted every 10ms, followed by frame stacking and subsampling~\cite{narayanan2019recognizing}. 
Specaugment~\cite{park2019specaugment} is used in the same manner as described in~\cite{park2020specaugment}. Both the first-pass and the deliberation models have a output vocabulary of 4,096 wordpiece units, including the end-of-sentence token.

For the first-pass RNN-T model, the encoder consists of 7 conformer layers with an attention dimension of 512, and for the first 3 layers we keep only the convolution operations to reduce on-device latency; the decoder uses a embedding prediction networks~\cite{botros2021tied} with dimension 640.
The first-pass model has a total of 56M weights.
It is separately trained with the RNN-T loss till convergence, and we freeze its parameters during second-pass model training.

\begin{table}[t]
    \caption{Architecture search for Align-Refine. We show WERs (\%) for different combinations of $(L,\, S)$ on the VS evaluation set, for up to 4 refinement steps during inference. First-pass RNN-T uses a beam size of $1$ for generating initial hypothesis, and has a WER of 8.4\%. Top: Fix $S$=3 and vary $L$. Bottom: Fix $L=6$ and vary $S$.}
    \label{tab:architecture-base}
    \centering
    \begin{tabular}{|@{\hspace{0.04\linewidth}}c@{\hspace{0.04\linewidth}}|@{\hspace{0.04\linewidth}}c@{\hspace{0.04\linewidth}}|r@{\hspace{0.04\linewidth}}|@{\hspace{0.04\linewidth}}r@{\hspace{0.04\linewidth}}|@{\hspace{0.04\linewidth}}r@{\hspace{0.04\linewidth}}|@{\hspace{0.04\linewidth}}r@{\hspace{0.04\linewidth}}|}
    \hline
        \multirow{2}{*}{$L$} & \multirow{2}{*}{$S$} & \multicolumn{4}{|c|}{Inference refinement step} \\ \cline{3-6}
        & & 1 & 2 & 3 & 4  \\
        \hline
        4 & 3 & 7.5 & 7.2 & 7.1 & 7.1 \\
        \textbf{6} & \textbf{3} & \textbf{6.9} & \textbf{6.5} & \textbf{6.5} & \textbf{6.4} \\
        $10$ & 3 & 7.3 & 7.0 & 6.9 & 6.9 \\ \hline
        6 & 1 & 7.5 & 10.0 & 47.8 & 63.8 \\
        6 & 2 & 7.0 & 6.6 & 6.6 & 6.5 \\
        6 & 4 & 7.0 & 6.5 & 6.5 & 6.4 \\ \hline
    \end{tabular}
\end{table}

\begin{table}[t]
    \caption{WERs (\%) of Align-Refine with proposed improvements on the VS evaluation set. Here $L=6$, $S=3$, and $D=512$. First-pass RNN-T uses a beam size of $1$ for generating initial hypothesis, and has a WER of 8.4\%.  Top: Cascaded encoder $enc_1$ with different number of conformer layers $L'$. Bottom: Alignment augmentation with different masking probability $p$. Note $(L',\,p)=(0,\,0)$ corresponds to best setup of Table~\ref{tab:architecture-base}. 
    }
    \label{tab:improvements}
    \centering
    \begin{tabular}{|@{\hspace{0.04\linewidth}}c@{\hspace{0.04\linewidth}}|@{\hspace{0.04\linewidth}}c@{\hspace{0.04\linewidth}}|r@{\hspace{0.04\linewidth}}|@{\hspace{0.04\linewidth}}r@{\hspace{0.04\linewidth}}|@{\hspace{0.04\linewidth}}r@{\hspace{0.04\linewidth}}|@{\hspace{0.04\linewidth}}r@{\hspace{0.04\linewidth}}|}
    \hline
        \multirow{2}{*}{$L'$} & \multirow{2}{*}{$p$} & \multicolumn{4}{|c|}{Inference refinement step} \\ \cline{3-6}
        & & 1 & 2 & 3 & 4  \\ \hline
        0 & 0 & 6.9 & 6.5 & 6.5 & 6.4 \\
        \hline
        4 & 0 & 6.5 & 6.1 & 6.1 & 6.1 \\
        5 & 0 & \textbf{6.4} & \textbf{6.1} & \textbf{6.0} & \textbf{6.0} \\ \hline \hline
        0 & 0.02 & \textbf{6.8} & \textbf{6.4} & \textbf{6.3} & \textbf{6.3} \\
        0 & 0.05 & 7.0 & 6.5 & 6.5 & 6.4 \\ 
        0 & 0.10 & 7.5 & 7.2 & 7.2 & 7.2 \\ 
        \hline
    \end{tabular}
\end{table}

\begin{table}[t]
    \caption{WERs (\%) of Align-Refine with increased first-pass decoding beam size. Here $L=6$, $S=3$, and $D=512$. First-pass RNN-T uses a beam size of $4$ for generating initial hypothesis, and has a WER of 7.8\%.}
    \label{tab:bs4-dev}
    \centering
    \begin{tabular}{|@{\hspace{0.04\linewidth}}c@{\hspace{0.04\linewidth}}|@{\hspace{0.04\linewidth}}c@{\hspace{0.04\linewidth}}|r@{\hspace{0.04\linewidth}}|@{\hspace{0.04\linewidth}}r@{\hspace{0.04\linewidth}}|@{\hspace{0.04\linewidth}}r@{\hspace{0.04\linewidth}}|@{\hspace{0.04\linewidth}}r@{\hspace{0.04\linewidth}}|}
    \hline
        \multirow{2}{*}{$L'$} & \multirow{2}{*}{$p$} & \multicolumn{4}{|c|}{Inference refinement step} \\ \cline{3-6}
        & & 1 & 2 & 3 & 4  \\ \hline
        0 & 0 & 6.7 & 6.4 & 6.4 & 6.4 \\
        4 & 0 & 6.2 & 5.9 & 5.9 & 5.9 \\
        4 & 0.02 & \textbf{6.1} &  \textbf{5.8} &  \textbf{5.7} &  \textbf{5.7} \\
        \hline
    \end{tabular}
\end{table}

\begin{figure*}[t]
    \centering
    \begin{tabular}{@{\hspace{0\linewidth}}|c|c|@{\hspace{0\linewidth}}} \hline
    First-pass RNN-T & Our method (step 4) \\ \hline\hline
\parbox{0.47\linewidth}{this {\color{blue} is} never happened in our state before and this should never be allowed said {\color{blue} when you go pal}} &  
\parbox{0.47\linewidth}{this {\color{red} has} never happened in our state before and this should never be allowed said {\color{red} venugopal}} \\[.6em] \hline

\parbox{0.47\linewidth}{ned evans denise thomas joe caffrey and incumbent christine {\color{blue} catsock one for your} terms on the board {\color{blue} and} tuesday's election} &  
\parbox{0.47\linewidth}{ned evans denise thomas joe caffrey and incumbent christine {\color{red} katsock won four-year} terms on the board {\color{red} in} tuesday's election} \\[.6em] \hline


\parbox{0.47\linewidth}{{\color{blue} are} churn rates have fallen back to levels {\color{blue} were} they were before {\color{blue} telling that} launched last year} &	
\parbox{0.47\linewidth}{{\color{red} our} churn rates have fallen back to levels {\color{red} where} they were before {\color{red} telenet} launched last year} \\[.6em] \hline

\parbox{0.47\linewidth}{on sunday the pair {\color{blue} we're seeing} enjoying another date night at the greenwich hotel {\color{blue} in} bowery hotel in new york} &
\parbox{0.47\linewidth}{on sunday the pair {\color{red} were seen} enjoying another date night at the greenwich hotel {\color{red} and} bowery hotel in new york} \\[.6em] \hline

\parbox{0.47\linewidth}{i've decided it's time to leave zynga {\color{blue} adventure} off on my own again said waldron}	& 
\parbox{0.47\linewidth}{i've decided it's time to leave zynga {\color{red} and venture} off on my own again said waldron}
\\[.6em] \hline
    \end{tabular}
    \vspace*{-.5em}
\caption{Case study of our method on the RPN-News evaluation set. 
    Blue color indicates errors and red indicates corrections.}
    \label{tab:case-study}
\end{figure*}

\subsection{Architecture search for Align-Refine}
\label{sec:architecture}

We first perform architecture search for the basic Align-Refine algorithm. Our $dec_1$ consists of $L$ transformer layers with attention dimension $D$ (and 8 attention heads), and we perform  $S$ refinement steps during training.

We set $D=512$ and search the hyperparameter combination $(L, S)$ based those used in the non-autoregressive decoding literature, and find $L=6$ and $S=3$ to provide a good trade off between final accuracy and training speed. 
In Table~\ref{tab:architecture-base} we provide sensitivity analysis for these parameters.
For this set of experiments, we set the beam size to 1 for the first-pass RNN-T model during both training and inference; the first-pass has a WER of 8.4\% on VS.
Observe that even with such a small beam size for the first-pass (which leads to fast inference), Align-Refine significantly improves its WER to 6.9\% in a single step, and 6.5\% with an additional step.

\subsection{Cascaded encoder and alignment augmentation}
\label{sec:improvements}

We then incorporate our proposed improvements to Align-Refine. 
With the architecture found from previous section, we explore cascaded encoders $enc_1$ with two different sizes. Our cascaded encoder consists of $L'$ conformer layers, each of which has a right context of $3$ encoder output frames, resulting in a total right context of $0.72$ seconds for $L'=4$, and $0.9$ seconds for $L'=5$. We found that both of them significantly improve over basic Align-Refine, while the benefit of using an additional layer with $L'=5$ is moderate, so we will be using $L'=4$ next in favor of a smaller model size.

Independent from using right context, we explore masking augmentation where we replace a percentage $p$ of the positions in $A^i$, $i=0,1,2$ with the special \texttt{[mask]} token during training. Dev set results for different $p$ values are provided in Table~\ref{tab:improvements}. We observe that too heavy masking ($p=0.10$) causes difficulty in training and degrades the final accuracy, while light augmentation ($p=0.02$) yields improvements.

Finally, we increase the beam size of first-pass RNN-T from $1$ to $4$ so as to obtain more accurate initial hypotheses 
(we still only use the top hypothesis and alignment for deliberation training), and retrain models without further hyperparameter tuning. The dev set results are shown in Table~\ref{tab:bs4-dev}. We observe that the first-pass model WER improves from 8.4\% to 7.8\%, and with the more accurate hypotheses, all deliberation setups achieved sizable WER improvement, while combining cascaded encoder and alignment augmentation yields the best accuracy, across all refinement steps.

In Figure~\ref{tab:case-study}, we provide examples where our method corrects first-pass decoding errors that sound similarly to the truth, showing that our method is able to capture context information from text and recover label dependency.

\subsection{Comparisons with other methods}
\label{sec:final_results}

We provide the test sets WERs of our base Align-Refine model (with $L=6$, $S=3$, $L'=0$, $p=0.0$) and best model (with $L=6$, $S=3$, $L'=4$, $p=0.02$, and first-pass beam size 4) in Table~\ref{tab:bs4-test}, along with comparisons with a few models.
Since Align-Refine uses the CTC loss for learning, we train a baseline CTC model that uses the same $enc_0$ (with weights taken from our fist-pass model and frozen during training) and an additional cascaded encoder of 5 conformer layers and 3 seconds right context; while this model sees large amount of right context, it does not model the label dependencies explicitly. 
We also compare with a second-pass RNN-T model that uses the same $enc_0$ (weights frozen during training) and an additional cascaded encoder with the same specifications of our $enc_1$; 
this is a baseline that explicitly models label dependency in the second-pass but does not look at first-pass hypotheses. 
Lastly, we compare with a deliberation model similar to that of~\cite{hu2021transformer},
where the decoder is a 512-dimensional 4-layer transformer decoder, and the text encoder consists of a 4-layer conformer with 3-token right context each layer. 
Except for our methods, others perform autoregressive decoding by beam search, with a beam size of $8$.

\begin{table}[t]
    \caption{Test set WERs (\%) of Align-Refine and baselines. We show WERs of our models at refinement step 4. The first-pass model has 56M weight parameters, and we provide the number of additional weights for each method.}
    \label{tab:bs4-test}
    \centering
    \begin{tabular}{|@{\hspace{0.01\linewidth}}c@{\hspace{0.01\linewidth}}|@{\hspace{0.01\linewidth}}c@{\hspace{0.01\linewidth}}|r@{\hspace{0.01\linewidth}}|@{\hspace{0.01\linewidth}}r@{\hspace{0.01\linewidth}}|@{\hspace{0.01\linewidth}}r@{\hspace{0.01\linewidth}}|@{\hspace{0.01\linewidth}}r@{\hspace{0.01\linewidth}}|@{\hspace{0.01\linewidth}}r@{\hspace{0.01\linewidth}}|@{\hspace{0.01\linewidth}}r@{\hspace{0.01\linewidth}}|}
    \hline
        \multirow{2}{*}{Method} & 2nd-pass &  \multicolumn{6}{c|}{WERs (\%)} \\ \cline{3-8}
        & \#weights & VS & SXS & RPN-M & -N & -P & -Q \\ \hline
        1st-pass & 0 & 7.8 & 37.5 & 16.6 & 11.4 & 40.9 & 25.6 \\
        Ours-base & 30M & 6.4 & 33.6 & 15.8 & 10.4 & 39.1 & 25.0 \\
        Ours-best & 55M & 5.7 & 32.0 & 14.6 & 10.0 & 38.3 & 23.5 \\
        \hline \hline
       2nd-pass CTC & 33M & 7.2 & 36.0 & 16.9 & 11.5 & 42.2 & 27.5 \\
       2nd-pass RNN-T & 35M & 5.8 & 31.8 & 13.9 & 9.6 & 38.7 & 22.1 \\
        Delib~\cite{hu2021transformer} & 48M& 6.0 & 34.3 & 13.8 & 10.2 & 36.2 & 22.2 \\
        \hline
    \end{tabular}
\end{table}

Observe that the basic Align-Refine method outperforms the second-pass CTC (even with a single step, see Table~\ref{tab:bs4-dev} first row for VS results), although the latter looks at even larger right context; this shows that conditioning on the first-pass hypotheses help capture much label dependency.
Our best model is on par with second-pass RNN-T and Delib on SXS, and worse on the rare word test sets. 
We leave it as future work to improve on rare word recognition. We emphasize that our method enjoys a much simpler and more efficient (parallelizable) inference procedure.


\section{Conclusions}
\label{sec:conclusions}
\vspace*{-.5em}

We have proposed a non-autoregressive decoding method for second-pass deliberation of a first-pass RNN-T. Our method improves the previously proposed Align-Refine algorithm by introducing cascaded encoder for the audio features, and  alignments augmentation by masking. We obtain significant WER reduction over the first-pass model with small amount of parameters. As future directions, we can extend our method to the streaming setup, and study techniques to improve accuracy on rare words.

\bibliographystyle{IEEEbib}
\pagebreak
\bibliography{refs}

\begin{thebibliography}{10}

\bibitem{xia2017deliberation}
Yingce Xia, Fei Tian, Lijun Wu, Jianxin Lin, Tao Qin, Nenghai Yu, and Tie-Yan
  Liu,
\newblock ``Deliberation networks: Sequence generation beyond one-pass
  decoding,''
\newblock in {\em Advances in Neural Information Processing Systems}, 2017, pp.
  1784--1794.

\bibitem{hu2020deliberation}
Ke~Hu, Tara~N. Sainath, Ruoming Pang, and Rohit Prabhavalkar,
\newblock ``Deliberation model based two-pass end-to-end speech recognition,''
\newblock in {\em 2020 IEEE International Conference on Acoustics, Speech and
  Signal Processing (ICASSP)}. IEEE, 2020, pp. 7799--7803.

\bibitem{hu2021transformer}
Ke~Hu, Ruoming Pang, Tara~N Sainath, and Trevor Strohman,
\newblock ``Transformer based deliberation for two-pass speech recognition,''
\newblock in {\em 2021 IEEE Spoken Language Technology Workshop (SLT)}. IEEE,
  2021, pp. 68--74.

\bibitem{sainath2020streaming}
Tara~N. Sainath, Yanzhang He, Bo~Li, Arun Narayanan, Ruoming Pang, Antoine
  Bruguier, Shuo-yiin Chang, Wei Li, Raziel Alvarez, Zhifeng Chen, et~al.,
\newblock ``A streaming on-device end-to-end model surpassing server-side
  conventional model quality and latency,''
\newblock in {\em 2020 IEEE International Conference on Acoustics, Speech and
  Signal Processing (ICASSP)}. IEEE, 2020, pp. 6059--6063.

\bibitem{ghazvininejad2019MaskPredict}
Yinhan Liu Luke~Zettlemoyer Marjan~Ghazvininejad, Omer~Levy,
\newblock ``Mask-predict: Parallel decoding of conditional masked language
  models,''
\newblock in {\em Proceedings of the 2019 Conference on Empirical Methods in
  Natural Language Processing}, 2019.

\bibitem{chen2019listen}
Nanxin Chen, Shinji Watanabe, Jes{\'u}s Villalba, and Najim Dehak,
\newblock ``Listen and fill in the missing letters: Non-autoregressive
  transformer for speech recognition,''
\newblock {\em arXiv preprint arXiv:1911.04908}, 2019.

\bibitem{higuchi2020mask}
Yosuke Higuchi, Shinji Watanabe, Nanxin Chen, Tetsuji Ogawa, and Tetsunori
  Kobayashi,
\newblock ``Mask ctc: Non-autoregressive end-to-end asr with ctc and mask
  predict,''
\newblock {\em arXiv preprint arXiv:2005.08700}, 2020.

\bibitem{higuchi2021improved}
Yosuke Higuchi, Hirofumi Inaguma, Shinji Watanabe, Tetsuji Ogawa, and Tetsunori
  Kobayashi,
\newblock ``Improved mask-ctc for non-autoregressive end-to-end asr,''
\newblock in {\em ICASSP 2021-2021 IEEE International Conference on Acoustics,
  Speech and Signal Processing (ICASSP)}, 2021.

\bibitem{wang2021streaming}
Tianzi Wang, Yuya Fujita, Xuankai Chang, and Shinji Watanabe,
\newblock ``Streaming end-to-end asr based on blockwise non-autoregressive
  models,''
\newblock {\em arXiv preprint arXiv:2107.09428}, 2021.

\bibitem{chan20}
William Chan, Chitwan Saharia, Geoffrey Hinton, Mohammad Norouzi, and Navdeep
  Jaitly,
\newblock ``Imputer: Sequence modelling via imputation and dynamic
  programming,''
\newblock in {\em Proceedings of the 37th International Conference on Machine
  Learning}.

\bibitem{chi2020align}
Ethan~A Chi, Julian Salazar, and Katrin Kirchhoff,
\newblock ``Align-refine: Non-autoregressive speech recognition via iterative
  realignment,''
\newblock {\em arXiv preprint arXiv:2010.14233}, 2020.

\bibitem{Chan_16a}
W.~Chan, N.~Jaitly, Q.~V. Le, and O.~Vinyals,
\newblock ``Listen, attend and spell: {A} neural network for large vocabulary
  conversational speech recognition,''
\newblock .

\bibitem{Graves_12a}
A.~Graves,
\newblock ``Sequence transduction with recurrent neural networks,''
\newblock in {\em ICML Workshop on Representation Learning}, 2012.

\bibitem{graves2006connectionist}
Alex Graves, Santiago Fern{\'a}ndez, Faustino Gomez, and J{\"u}rgen
  Schmidhuber,
\newblock ``Connectionist temporal classification: {Labelling} unsegmented
  sequence data with recurrent neural networks,''
\newblock in {\em Proceedings of the 23rd international conference on Machine
  learning}, 2006.

\bibitem{He_18a}
Y.~He, T.~Sainath, R.~Prabhavalkar, I.~McGraw, and \textit{et al.},
\newblock ``Streaming end-to-end speech recognition for mobile devices,''
\newblock in {\em ICASSP}, 2019.

\bibitem{Vaswani_17a}
Ashish Vaswani, Noam Shazeer, Niki Parmar, Jakob Uszkoreit, Llion Jones,
  Aidan~N Gomez, \L~ukasz Kaiser, and Illia Polosukhin,
\newblock ``Attention is all you need,''
\newblock in {\em Advances in Neural Information Processing Systems}, I.~Guyon,
  U.~V. Luxburg, S.~Bengio, H.~Wallach, R.~Fergus, S.~Vishwanathan, and
  R.~Garnett, Eds. 2017, Curran Associates, Inc.

\bibitem{arun21cascade}
A.~Narayanan, T.~N. Sainath, R.~Pang, et~al.,
\newblock ``{Cascaded encoders for unifying streaming and non-streaming ASR},''
\newblock in {\em Proc. ICASSP}, 2021.

\bibitem{gulati2020conformer}
A.~Gulati, J.~Qin, C.-C. Chiu, et~al.,
\newblock ``{Conformer: Convolution-augmented Transformer for Speech
  Recognition},''
\newblock in {\em Proc. Interspeech}, 2020.

\bibitem{devlin2018bert}
Jacob Devlin, Ming-Wei Chang, Kenton Lee, and Kristina Toutanova,
\newblock ``Bert: Pre-training of deep bidirectional transformers for language
  understanding,''
\newblock {\em arXiv preprint arXiv:1810.04805}, 2018.

\bibitem{park2019specaugment}
Daniel~S Park, William Chan, Yu~Zhang, Chung-Cheng Chiu, Barret Zoph, Ekin~D
  Cubuk, and Quoc~V Le,
\newblock ``Specaugment: A simple data augmentation method for automatic speech
  recognition,''
\newblock {\em arXiv preprint arXiv:1904.08779}, 2019.

\bibitem{narayanan2019recognizing}
Arun Narayanan, Rohit Prabhavalkar, Chung-Cheng Chiu, David Rybach, Tara~N
  Sainath, and Trevor Strohman,
\newblock ``Recognizing long-form speech using streaming end-to-end models,''
\newblock in {\em ASRU}, 2019.

\bibitem{kim2017generation}
Chanwoo Kim, Ananya Misra, Kean Chin, Thad Hughes, Arun Narayanan, Tara
  Sainath, and Michiel Bacchiani,
\newblock ``Generation of large-scale simulated utterances in virtual rooms to
  train deep-neural networks for far-field speech recognition in google home,''
\newblock 2017.

\bibitem{li2012improving}
Jinyu Li, Dong Yu, Jui-Ting Huang, and Yifan Gong,
\newblock ``Improving wideband speech recognition using mixed-bandwidth
  training data in cd-dnn-hmm,''
\newblock in {\em 2012 IEEE Spoken Language Technology Workshop (SLT)}.

\bibitem{Golan16}
G.~Pundak and T.~N. Sainath,
\newblock ``Lower frame rate neural network acoustic models,''
\newblock in {\em Proc. Interspeech}, 2016.

\bibitem{park2020specaugment}
Daniel~S Park, Yu~Zhang, Chung-Cheng Chiu, Youzheng Chen, Bo~Li, William Chan,
  Quoc~V Le, and Yonghui Wu,
\newblock ``Specaugment on large scale datasets,''
\newblock in {\em ICASSP 2020-2020 IEEE International Conference on Acoustics,
  Speech and Signal Processing (ICASSP)}. IEEE, 2020, pp. 6879--6883.

\bibitem{botros2021tied}
Rami Botros, Tara~N Sainath, Robert David, Emmanuel Guzman, Wei Li, and
  Yanzhang He,
\newblock ``Tied \& reduced rnn-t decoder,''
\newblock {\em arXiv preprint arXiv:2109.07513}, 2021.

\end{thebibliography}

\end{document}